\documentclass[runningheads]{llncs}
\usepackage{graphicx}
\usepackage{amsmath}
\usepackage{comment}

\newcommand\blfootnote[1]{%
  \begingroup
  \renewcommand\thefootnote{}\footnote{#1}%
  \addtocounter{footnote}{-1}%
  \endgroup
}

\begin{document}
\title{Using Transformer based Ensemble Learning to classify Scientific Articles}
%FideLIPI @ SDPRA 2021: 
%\thanks{Supported by Fidelity Investments.}}
\author{Sohom Ghosh\thanks{Equal Contribution} \orcidID{0000-0002-4113-0958}\and
Ankush Chopra\thanks{Equal Contribution} \orcidID{0000-0002-9970-8038}}
\institute{
Artificial Intelligence, CoE, Fidelity Investments, Bengaluru, Karnataka, India \\
\email{\{sohom1ghosh,ankush01729\}@gmail.com}}
\maketitle              % typeset the header of the contribution
\begin{abstract}
Many time reviewers fail to appreciate novel ideas of a researcher and provide generic feedback. Thus, proper assignment of reviewers based on their area of expertise is necessary. Moreover, reading each and every paper from end-to-end for assigning it to a reviewer is a tedious task. In this paper, we describe a system which our team FideLIPI submitted in the shared task of SDPRA-2021\footnote{https://sdpra-2021.github.io/website/ (accessed January 25, 2021)} \cite{reddysdpra2021}. It comprises four independent sub-systems capable of classifying abstracts of scientific literature to one of the given seven classes. The first one is a RoBERTa \cite{liu2019roberta} based model built over these abstracts. Adding topic models / Latent dirichlet allocation (LDA) \cite{10.5555/944919.944937} based features to the first model results in the second sub-system. The third one is a sentence level RoBERTa \cite{liu2019roberta} model. The fourth one is a Logistic Regression model built using Term Frequency Inverse Document Frequency (TF-IDF) features. We ensemble predictions of these four sub-systems using majority voting to develop the final system which gives a F1 score of 0.93 on the test and validation set. This outperforms the existing State Of The Art (SOTA) model SciBERT's \cite{beltagy-etal-2019-scibert} in terms of F1 score on the validation set. Our codebase is available at \url{https://github.com/SDPRA-2021/shared-task/tree/main/FideLIPI}
%abstract up to 15-200 words

\end{abstract}
%
%
%System description papers should not be more than 8 pages.
\section{Introduction}
\blfootnote{In: Gupta M., Ramakrishnan G. (eds) Trends and Applications in Knowledge Discovery and Data Mining. PAKDD 2021. Lecture Notes in Computer Science, vol 12705. Springer, Cham. https://doi.org/10.1007/978-3-030-75015-2\_11 \\ This is the author's version.}
Due to the ever-increasing number of research paper submissions per conference, it has become extremely difficult to manually assign appropriate reviewers to a paper based on their expertise. Improper assignment of reviewer leads to poor quality of reviews. Thus, an automated system capable of determining which category a research paper belongs to is necessary to develop. A similar shared task has been hosted at The First Workshop \& Shared Task on Scope Detection of the Peer Review Articles (SDPRA-2021) (Collocated with the 25\textsuperscript{th} Pacific-Asia Conference on Knowledge Discovery and Data Mining\footnote{https://www.pakdd2021.org/ (accessed January 25, 2021)} (PAKDD-2021)) \cite{reddysdpra2021}. This paper narrates the approach our team FideLIPI followed while participating in this challenge. It is structured into five main sections. This section introduces readers to the problem that we have tried to solve, related works that have been done previously and our contributions. The next section familiarizes readers with the dataset, pre-processing and feature engineering steps. The section following it describes the system we developed and how we generated the submissions we have made. The subsequent section discusses the experiments we performed and their results. After that, we conclude the paper by revealing our plans to improve the system further.

\subsection{Problem Description}
Given an abstract A of a scientific article, we need to assign one of the seven classes (CL, CR, DC, DS, LO, NI, SE)\footnote{These abbreviations have been expanded in Table \ref{tab:data-description}} to it.

\subsection{Related Works}
Beltagy et al. in their paper \cite{beltagy-etal-2019-scibert} propose SciBERT which is a language model trained specifically for scientific articles. They outperformed the BERT \cite{devlin-etal-2019-bert} base model in domains like Bio-medicine and Computer Science. In the paper \cite{cao05} Cao et al. describe how contents as well as citations can be used to classify scientific documents. They try out different Machine Learning models like K Nearest Neighbours, Nearest Centroid and Naive Bayes. One of the pioneering work has been done by Ghanem et al. \cite{ghanem}. While participating in KDD CUP 2002 (Task 1) they proposed a novel method for extracting frequently appearing keywords. They use these patterns as input to a Support Vector Machine (SVM) classifier \cite{svm}. Borrajo et al.  \cite{borrajo} studied how over-sampling, under-sampling, use of dictionaries  help in classifying scientific articles relating to bio-medicine. They mainly dealt with three kinds of classifiers: K Nearest Neighbour, SVM and Naive-Bayes. For evaluation, they used Precision, Recall, F-measure and Utility. They achieved best results by using sub-sampling along with NLPBA, Protein dictionaries and a SVM classifier \cite{svm}.

%\begin{comment}
\subsection{Our Contributions}
Our contributions are as follows:
 \begin{itemize}
    \item We have developed a system capable of assigning a class to the abstract of a given scientific article. Our model (F1: 0.928) surpassed the existing SOTA model SciBERT's \cite{beltagy-etal-2019-scibert} performance (F1: 0.926) in the given validation set.
     \item For enhancing reproducibility and transparency, we have open-sourced the system. It is available here\footnote{https://github.com/SDPRA-2021/shared-task/tree/main/FideLIPI}.
 \end{itemize}
%\end{comment}

\section{Dataset}
This section narrates in details the data we are dealing with and the pre-processing steps we followed.

\subsection{Data Description}
The dataset \cite{sainireddy2021} for the shared task of The First Workshop and Shared Task on Scope Detection of the Peer Review Articles (SPDRA 2021) \cite{reddysdpra2021} consists of 16,800 training instances, 11,200 validation instances and 7,000 test instances. They belong to seven classes as described in Table \ref{tab:data-description}.

\begin{table}%[]
\centering
\caption{Data Description }
\label{tab:data-description}
\begin{tabular}{|l|l|l|l|}
\hline
\textbf{Category}                                                & \textbf{Train} & \textbf{Validation} & \textbf{Test} \\ \hline
{ Computation and Language (CL)}             & 2,740           & 1,866                & 1,194          \\ \hline
{ Cryptography and Security (CR)}            & 2,660           & 1,835                & 1,105          \\ \hline
{ Distributed and Cluster Computing (DC)}    & 2,042           & 1,355                & 803           \\ \hline
{ Data Structures and Algorithms (DS)}       & 2,737           & 1,774                & 1,089          \\ \hline
{ Logic in Computer Science (LO)}            & 1,811           & 1,217                & 772           \\ \hline
{ Networking and Internet Architecture (NI)} & 2,764           & 1,826                & 1,210          \\ \hline
{ Software Engineering (SE)}                 & 2,046           & 1,327                & 827           \\ \hline
\textbf{TOTAL}                                                   & \textbf{16,800} & \textbf{11,200}      & \textbf{7,000} \\ \hline
\end{tabular}
\end{table}

\subsection{Pre-processing \& Feature Engineering}
For Model-1 narrated in section \ref{model-1}, we keep the raw abstracts as it is and we do not pre-process them. For Model-2, after converting the abstracts to lowercase, removing stop words and lemmatizing them, we empirically decide to extract 50 topics using Topic Modelling (LDA) \cite{10.5555/944919.944937}. This is narrated in section \ref{model-2}. We used NLTK\footnote{https://www.nltk.org/ (accessed January 25, 2021)} and Gensim\footnote{https://radimrehurek.com/gensim/ (accessed January 25, 2021)} libraries to achieve this.
As pre-processing steps of Model-3 described in section \ref{model-3}, we remove newline characters and extract sentences from abstracts of scientific articles using Spacy\footnote{https://spacy.io/ (accessed January 25, 2021)} library. For Model-4 mentioned in section \ref{model-4}, we remove stop words and create TF-IDF based features with n-grams ranging from 1 to 4. We further ignore the terms with document frequency strictly lesser than 0.0005. We selected these hyper-parameters through experimentation. This resulted in 22,151 features.

\section{Methodology}
In this section, we describe each of the sub-systems/models which are ensembled to create the final system. Moreover, we elucidate the process we followed to generate each of the three submissions we have made.\\
NOTE: All of the hyper-parameters mentioned in this section have been obtained through rigorous experimentation described in section \ref{expsetup} . The hyper-parameters of the RoBERTa \cite{liu2019roberta} models used here  are mentioned in Table \ref{tab:hyperparameters}.

\subsection{System Description of Model-1 (RoBERTa)}
\label{model-1}
It is a RoBERTa \cite{liu2019roberta} based model built on the raw text corpus of the abstract. Its task is to predict probability of each of the given classes. We finally select the class having maximum probability.

\subsection{System Description of Model-2 (RoBERTa+LDA)}
\label{model-2}
This model is similar to the one described in section \ref{model-1}. It additionally takes 50 Topic Modelling (LDA) \cite{10.5555/944919.944937} based features. Additional dropout of 0.3 is implemented in the classification layer of this RoBERTa \cite{liu2019roberta} model.

\subsection{System Description of Model-3 (RoBERTa on sentences)}
\label{model-3}
In this model, instead of considering the whole abstract as input, we split it into individual sentences. Sentences having a number of tokens greater or equal to ten are considered while training the RoBERTa \cite{liu2019roberta} based model. While scoring the validation set we considered only those sentences which have a number of tokens greater or equal to six. These numbers have been obtained empirically.  To decide the final label of a given abstract, we add the logarithmic probabilities of predictions of the individual sentences for each of the seven classes. We then, select the class for which this value is maximum.

\subsection{System Description of Model-4 (TF-IDF + Logistic Regression)}
\label{model-4}
This is a simple logistic regression model built using scikit-learn\footnote{https://scikit-learn.org/ (accessed January 25, 2021)} library over 22,151 TF-IDF features. Its hyper-parameters are as follows:
 maximum number of iterations = 100,  penalty = l2, tolerance = 0.0001.

\subsection{Submissions}
\label{submission}
\begin{figure}
\includegraphics[width=\textwidth]{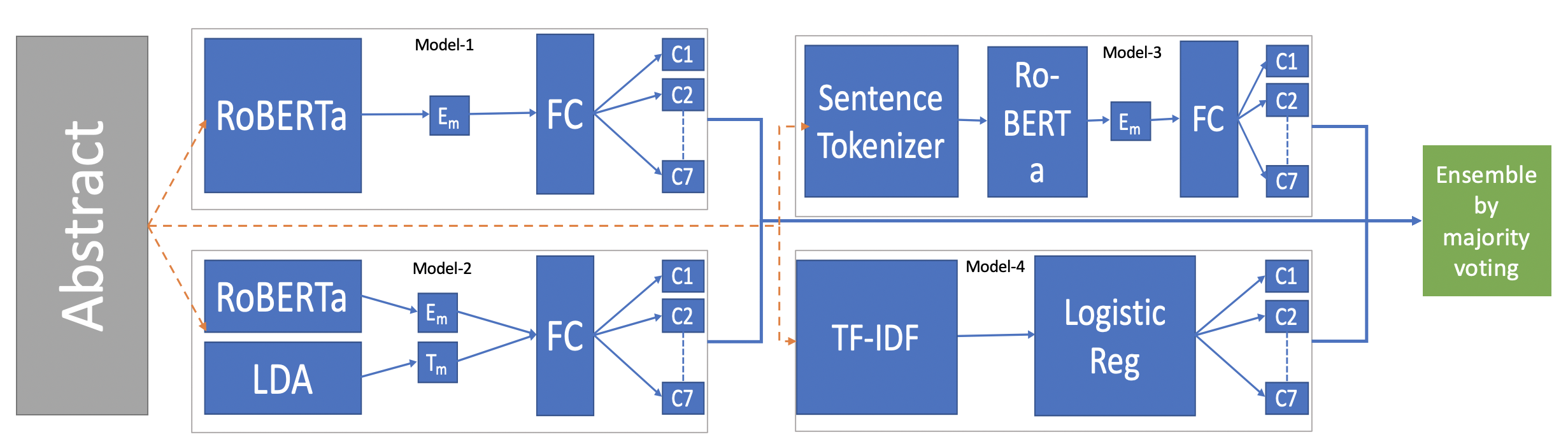}
\caption{Ensemble Model Architecture. E\textsubscript{m} = Embedding, T\textsubscript{m} = Topic Models, FC = Fully Connected Layer, C\textsubscript{1} to C\textsubscript{2} are the classes corresponding to CL, CR and so on} \label{fig:blockdiagram}
\end{figure}

As per the rules of this shared task, each team could submit at-most three sets of predictions for the test set. Our first submission is an ensemble of all the four models described above using majority voting technique. It has been depicted in Figure \ref{fig:blockdiagram}. In this technique, the class which gets the maximum number of vote is selected as the final class. Whenever there is a  tie between two classes, we randomly choose one of them. The second and third submissions are results of Model-2 (refer to section \ref{model-2}) and Model-4 (refer to section \ref{model-4}) respectively.

%Ensemble Model 1-2-3-4 using 
\section{Experiments, Results and Discussion}
\subsection{Experimental Setup}
\label{expsetup}
We started with zero-shot learning \cite{Chang2008ImportanceOSzeroshot} and pre-trained models like BERT \cite{devlin-etal-2019-bert}, RoBERTa \cite{liu2019roberta} and T5 \cite{2020t5}, since the transformer-based pre-trained models are producing state-of-the-art results on most of the Natural Language Processing (NLP) tasks. We used text classification module of the Simple Transformer library\footnote{https://simpletransformers.ai/ (accessed January 25, 2021)} to run the multi-class classification using the BERT [\cite{devlin-etal-2019-bert} model. The Simple Transformer library provides an easy interface to run different NLP tasks while using the HuggingFace \cite{wolf-etal-2020-transformers} Transformers at the back-end. We fine-tune the underlying pre-trained model while training it for the task. We ran classification using BERT-base \cite{devlin-etal-2019-bert} for 10 epochs and saved model after each epoch. Based on the saturation of validation set performance improvement, we chose the right epoch for both the models.

T5 \cite{2020t5} uses both encoder and decoder parts of the transformer. Although both input and output of the model need to be text sequence, it can still be used effectively for the text classification task. We utilized the T5 Model \cite{2020t5} class of Simple Transformer to train a model in multi-class classification setting using T5-base \cite{2020t5} pre-trained model. We used the HuggingFace \cite{wolf-etal-2020-transformers} library for training a classifier using the pre-trained RoBERTa \cite{liu2019roberta} model. We passed the abstracts through the RoBERTa \cite{liu2019roberta} model and took the embedding of the [CLS] token which was passed through a classification head set up to train for multi-class classification.

We decided to perform further experimentation with the RoBERTa \cite{liu2019roberta} model since it had the best performance among the three vanilla model that we built. We created a representation of the abstracts using TF-IDF, and LDA \cite{10.5555/944919.944937} based topic modelling techniques. We combined the LDA  \cite{10.5555/944919.944937} and TF-IDF features with the RoBERTa \cite{liu2019roberta} individually and trained two models. We could see the slight improvement in the performance when we combined LDA \cite{10.5555/944919.944937} with the RoBERTa \cite{liu2019roberta} compared to vanilla RoBERTa \cite{liu2019roberta}  model. RoBERTa \cite{liu2019roberta} has a limitation of 512 tokens as input. Many of the abstracts have more than 512 tokens. Thus, the part towards the tail of the abstract was not getting utilized for prediction in these cases. So, we decided to train a sentence level model, where we first tokenized the abstract into sentences using Spacy. These tokenized sentences were assigned the same token as the abstract they were part of. We then trained a classification model using this data using Simple Transformer library with RoBERTa \cite{liu2019roberta} base model. We observed that very short sentences seldom carry enough information to be able to predict the right class just using their constituent words. Hence, we built four models first three by not considering a sentence that had less than 1, 6 and 10 words and the fourth one by taking all the sentences. We found that the model where we had taken sentences with more than 10 words did better than all other models. While scoring these models on the validation data, we scored the individual sentence and averaged the output probability for all the sentences from an abstract. Then, the class with the highest average probability was assigned as the prediction for the abstract. We experimented with the length during the predictions as well and found that taking sentences with more than 5 words tend to perform the best.
Finally, we bench-marked our model with one an existing SOTA model which is SciBERT \cite{beltagy-etal-2019-scibert}.

\subsection{Results and Discussion}
The results\footnote{NOTE: The F1 scores corresponding to the test set have been provided by the SDPRA team after evaluating three of our submissions. Since the number of submissions was restricted to three, we do not have these numbers for the other models which we have developed.} we obtained are presented in Table \ref{tab:results}. Here F1 refers to the weighted F1 score. Details relating to models 1,2,3 and 4 have been mentioned in the previous section of this paper.
\begin{table}%[]
\centering
\caption{Performance of various models. \textbf{Bold} highlights the best performing models. \underline{Underline} denotes existing State Of The Art (SOTA) model.}
\label{tab:results}
\begin{tabular}{|l|l|l|l|}
\hline
\textbf{Model}           & \textbf{Train (F1)} & \textbf{Validation (F1)} & \multicolumn{1}{l|}{\textbf{Test (F1)}} \\ \hline
Zero shot learning & 0.114 & 0.113 & \_                          \\ \hline
BERT              &  \textbf{0.999}     &  0.898     & \_                          \\ \hline
T5                &  0.977     &  0.882     & \_                          \\ \hline
\underline{SciBERT (SOTA)}          & \underline{0.991}      & \underline{0.926} & \_                          \\ \hline
Model-1 (RoBERTa)          & \textbf{0.999} & 0.916 & \_                          \\ \hline
Model-2 (RoBERTa+LDA)         & 0.968 & 0.919 & \multicolumn{1}{l|}{0.912} \\ \hline
Model-3 (RoBERTa on sentences)          & \textbf{0.999} & 0.913 & \_                          \\ \hline
Model-4 (TF-IDF+Logistic Regression)          & 0.958  & 0.915  & \multicolumn{1}{l|}{0.916} \\ \hline
\textbf{Ensemble (Model 1-2-3-4)} & \textbf{0.999}               & \textbf{0.928}                    & \multicolumn{1}{l|}{\textbf{0.929}}             \\ \hline
\end{tabular}
\end{table}

\begin{table}
\centering
\caption{Hyper-parameters of various models}
\label{tab:hyperparameters}
\begin{tabular}{|l|l|l|l|l|l|l|}
\hline
\textbf{Model} & \textbf{Max. Token} & \textbf{BS (Train)} & \textbf{BS (Valid)} & \textbf{\# Epochs} & \textbf{$\eta$} & \textbf{Optimizer} \\ \hline
RoBERTa-1    & 512 & 32  & 256   & 13 & 0.00002 &ADAM\cite{kingma2017adam} \\ \hline
RoBERTa-2   & 512 & 32  & 8   & 3 & 0.00002 & ADAM\cite{kingma2017adam} \\ \hline
RoBERTa-3   & 512 & 8  & 8   & 10 & 0.00004 & AdamW\cite{adamw} \\ \hline
BERT    & 512 & 8  & 8   & 10 & 0.00004 & AdamW\cite{adamw} \\ \hline
T5      & 512 & 8  & 8   & 7  & 0.00100 & ADAM\cite{kingma2017adam}  \\ \hline
SciBERT & 512 & 32 & 256 & 5  & 0.00002 & ADAM\cite{kingma2017adam}  \\ \hline
\end{tabular}
\end{table}

The hyper-parameters corresponding to the best performing versions of RoBERTa \cite{liu2019roberta} models described in sections \ref{model-1} (RoBERTa-1), \ref{model-2} (RoBERTa-2), \ref{model-3} (RoBERTa-3), BERT \cite{devlin-etal-2019-bert} model, T5 \cite{2020t5} model and SciBERT \cite{beltagy-etal-2019-scibert} model are mentioned in Table \ref{tab:hyperparameters}. Max. Token is the maximum number of tokens, BS means the batch size and $\eta$ represents the learning rate.

%\subsection{Discussion}
While performing the experiments, we observed that the Logistic Regression model takes the least amount of training time as compared to other transformer-based models. In each of these models, we saw that the performance was worst for the class Distributed and Cluster Computing (DC) and it was best for the class Computation and Language (CL). On analysing Table \ref{tab:results} we see that the F1 scores are 0.999 on the training set for BERT \cite{devlin-etal-2019-bert}, Model-1, Model-3 and the ensemble model. However, the ensemble model performs best for Validation set (F1 = 0.928). We further observe that zero-shot learning (F1=0.114 on the training set and F1=0.113 on the validation set) performs worse than all other models. This is because it has not seen the training data. This also confirms that generic models do not perform well on close-ended domain-specific tasks.

\section{Conclusion and Future Works}
Analysing the results mentioned in the previous section, we conclude that the individual models' performances are comparable. We further observe that the ensembling technique outperforms existing SOTA model SciBERT \cite{beltagy-etal-2019-scibert} on the validation set in terms of F1.

In future, we would like to experiment by replacing the RoBERTa \cite{liu2019roberta} based embeddings with SciBERT \cite{beltagy-etal-2019-scibert} in the sub-systems mentioned in sections \ref{model-1}, \ref{model-2} and \ref{model-3}. Furthermore, we want to study how replacing the majority voting method of section \ref{submission} with a meta classifier or a fully connected dense layer affects the overall performance. Finally, we shall be working on reducing the model size.

 %\bibliographystyle{splncs04}
 %\bibliography{mybibliography}
 
%

\bibliographystyle{splncs04}
\bibliography{references/reddy-saini-sdpra21-dataset.bib, references/D19-1371.bib, references/roberta.bib, references/acm_944919.944937.bib, references/225524714.bib, references/S0957417416301464.bib, references/S095741741830215X.bib, references/cao.bib, references/citation-321660141.bib, references/acm_772862_772876.bib, references/adam.bib, references/bert.bib, references/t5.bib, references/huggingface.bib, references/zeroshot.bib, references/svm.bib, references/adamw.bib, references/reddysdpra.bib}  

\begin{thebibliography}{10}
\providecommand{\url}[1]{\texttt{#1}}
\providecommand{\urlprefix}{URL }
\providecommand{\doi}[1]{https://doi.org/#1}

\bibitem{beltagy-etal-2019-scibert}
Beltagy, I., Lo, K., Cohan, A.: {S}ci{BERT}: A pretrained language model for
  scientific text. In: Proceedings of the 2019 Conference on Empirical Methods
  in Natural Language Processing and the 9th International Joint Conference on
  Natural Language Processing (EMNLP-IJCNLP). pp. 3615--3620. Association for
  Computational Linguistics, Hong Kong, China (Nov 2019).
  \doi{10.18653/v1/D19-1371}, \url{https://www.aclweb.org/anthology/D19-1371}

\bibitem{10.5555/944919.944937}
Blei, D.M., Ng, A.Y., Jordan, M.I.: Latent dirichlet allocation. J. Mach.
  Learn. Res.  \textbf{3}(null),  993–1022 (Mar 2003)

\bibitem{borrajo}
Borrajo, M., Romero, R., Iglesias, E., Marey, C.: Improving imbalanced
  scientific text classification using sampling strategies and dictionaries.
  Journal of Integrative Bioinformatics  \textbf{8} (12 2011).
  \doi{10.1515/jib-2011-176}

\bibitem{cao05}
Cao, M.D., Gao, X.: Combining contents and citations for scientific document
  classification. In: Zhang, S., Jarvis, R. (eds.) AI 2005: Advances in
  Artificial Intelligence. pp. 143--152. Springer Berlin Heidelberg, Berlin,
  Heidelberg (2005)

\bibitem{Chang2008ImportanceOSzeroshot}
Chang, M.W., Ratinov, L.A., Roth, D., Srikumar, V.: Importance of semantic
  representation: Dataless classification. In: AAAI (2008)

\bibitem{svm}
Cortes, C., Vapnik, V.: Support-vector networks. Machine learning
  \textbf{20}(3),  273--297 (1995)

\bibitem{devlin-etal-2019-bert}
Devlin, J., Chang, M.W., Lee, K., Toutanova, K.: {BERT}: Pre-training of deep
  bidirectional transformers for language understanding. In: Proceedings of the
  2019 Conference of the North {A}merican Chapter of the Association for
  Computational Linguistics: Human Language Technologies, Volume 1 (Long and
  Short Papers). pp. 4171--4186. Association for Computational Linguistics,
  Minneapolis, Minnesota (Jun 2019). \doi{10.18653/v1/N19-1423},
  \url{https://www.aclweb.org/anthology/N19-1423}

\bibitem{ghanem}
Ghanem, M.M., Guo, Y., Lodhi, H., Zhang, Y.: Automatic scientific text
  classification using local patterns: Kdd cup 2002 (task 1). SIGKDD Explor.
  Newsl.  \textbf{4}(2),  95–96 (Dec 2002). \doi{10.1145/772862.772876},
  \url{https://doi.org/10.1145/772862.772876}

\bibitem{kingma2017adam}
Kingma, D.P., Ba, J.: Adam: A method for stochastic optimization (2017)

\bibitem{liu2019roberta}
Liu, Y., Ott, M., Goyal, N., Du, J., Joshi, M., Chen, D., Levy, O., Lewis, M.,
  Zettlemoyer, L., Stoyanov, V.: Roberta: A robustly optimized bert pretraining
  approach (2019), \url{http://arxiv.org/abs/1907.11692}, cite arxiv:1907.11692

\bibitem{adamw}
Loshchilov, I., Hutter, F.: Fixing weight decay regularization in adam. CoRR
  \textbf{abs/1711.05101} (2017), \url{http://arxiv.org/abs/1711.05101}

\bibitem{2020t5}
Raffel, C., Shazeer, N., Roberts, A., Lee, K., Narang, S., Matena, M., Zhou,
  Y., Li, W., Liu, P.J.: Exploring the limits of transfer learning with a
  unified text-to-text transformer. Journal of Machine Learning Research
  \textbf{21}(140),  1--67 (2020), \url{http://jmlr.org/papers/v21/20-074.html}

\bibitem{sainireddy2021}
Reddy, Saichethan;~Saini, N.: ``sdpra 2021 shared task data”, mendeley data,
  v1, (2021). \doi{10.17632/njb74czv49.1},
  \url{https://data.mendeley.com/datasets/njb74czv49/1}

\bibitem{reddysdpra2021}
Reddy, S., Saini., N.: Overview and insights from scope detection of the peer
  review articles shared tasks 2021 (forthcoming). In: Proceedings of The First
  Workshop \& Shared Task on Scope Detection of the Peer Review Articles (SDPRA
  2021) (2021)

\bibitem{wolf-etal-2020-transformers}
Wolf, T., Debut, L., Sanh, V., Chaumond, J., Delangue, C., Moi, A., Cistac, P.,
  Rault, T., Louf, R., Funtowicz, M., Davison, J., Shleifer, S., von Platen,
  P., Ma, C., Jernite, Y., Plu, J., Xu, C., Scao, T.L., Gugger, S., Drame, M.,
  Lhoest, Q., Rush, A.M.: Transformers: State-of-the-art natural language
  processing. In: Proceedings of the 2020 Conference on Empirical Methods in
  Natural Language Processing: System Demonstrations. pp. 38--45. Association
  for Computational Linguistics, Online (Oct 2020),
  \url{https://www.aclweb.org/anthology/2020.emnlp-demos.6}

\end{thebibliography}
% RoBERTa, LDA, TF-IDF






% LNCS Homepage, \url{http://www.springer.com/lncs}. Last accessed 4

% \end{thebibliography}

\end{document}